\documentclass[letterpaper, 10 pt, conference]{ieeeconf}  
\IEEEoverridecommandlockouts
\usepackage[ruled]{algorithm2e}
\usepackage{mathrsfs,amsmath} 
\usepackage{amssymb}  
\usepackage{graphicx}
\usepackage{bm}
\usepackage{cite}
\usepackage{gensymb}
\usepackage{multirow}
\usepackage{threeparttable}
\usepackage{hyperref}
\usepackage{subcaption}
\usepackage{booktabs} 

\usepackage{color}

\title{\LARGE \bf
Optimal Robotic Velcro Peeling with Force Feedback
}

\author{Jiacheng Yuan$^{1}$, Changhyun Choi$^{1}$ and Volkan Isler$^{2}$
    \thanks{$^{1}$ are with Department of Electrical and Computer Engineering, University of Minnesota, Minneapolis, MN, 55455, USA {\tt\small yuanx320, cchoi@umn.edu}}%
    \thanks{$^{2}$ is with the Department of Computer Science and Engineering, University of Minnesota, Minneapolis, MN, 55455, USA {\tt\small isler@umn.edu}}%
}

\begin{document}
\maketitle
\thispagestyle{empty}
\pagestyle{empty}
\setlength{\parskip}{0pt}

\begin{abstract}
We study the problem of peeling a Velcro strap from a surface using a robotic manipulator. The surface geometry is arbitrary and unknown. The robot has access to only the force feedback and its end-effector position. This problem is challenging due to the partial observability of the environment and the incompleteness of the sensor feedback. To solve it, we first model the system with simple analytic state and action models based on quasi-static dynamics assumptions. We then study the fully-observable case where the state of both the Velcro and the robot are given. For this case, we obtain the optimal solution in closed-form which minimizes the total energy cost. Next, for the partially-observable case, we design a state estimator which estimates the underlying state using only force and position feedback. Then, we present a heuristics-based controller that balances exploratory and exploitative behaviors in order to peel the velcro efficiently. Finally, we evaluate our proposed method in environments with complex geometric uncertainties and sensor noises, achieving 100\% success rate with less than 80\% increase in energy cost compared to the optimal solution when the environment is fully-observable, outperforming the baselines by a large margin.


\end{abstract}
\section{Introduction}

\begin{figure}[t]
\centering
\includegraphics[width=0.4\textwidth]{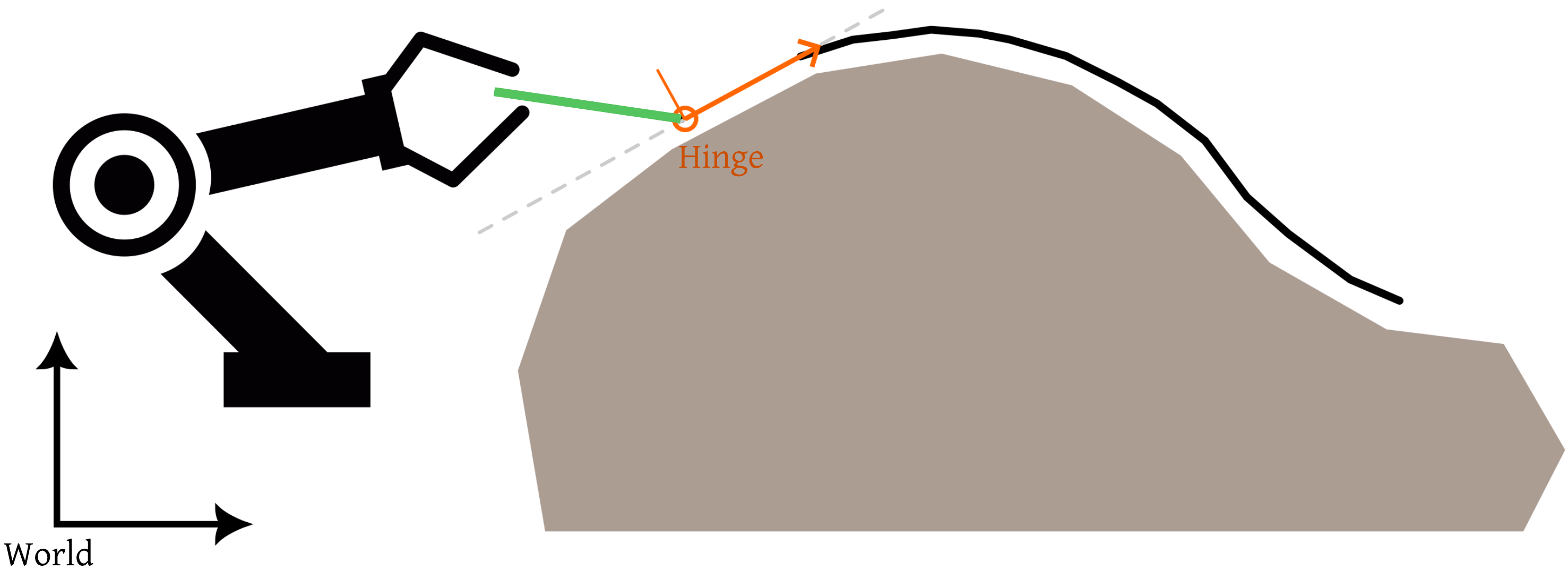}
\caption{\small Velcro peeling diagram illustrating a robotic manipulator grasping the tip of the peeled part (green) of the Velcro strap and the rest of Velcro is applied on a rigid object with a varying contour. Hinge refers to the connecting point of the peeled and attached part with a vector pointing the orientation of the following segments.}
\label{fig:main_figure}
\vspace{-7mm}
\end{figure}

Robots are moving beyond the confines of controlled factory settings and into our homes, workplaces, and public spaces where they are expected to operate in unstructured and uncertain environments.
Such transition requires a significant increase in their ability to perceive and manipulate objects within environments where complete knowledge of the surroundings is often unavailable or sensor observations have ambiguity. Furthermore, they need to handle a broad set of objects, including deformable objects such as shoes which might have velcro straps or laces. This work delves into a challenging problem which arises in such settings:   peeling Velcro straps applied on unknown geometry while relying solely on force and proprioceptive feedback (Fig.~\ref{fig:main_figure}).

The Velcro peeling task arises, for example,  when assisting a person with putting on or taking off shoes and coats, or opening grocery delivery bags. 
It encapsulates the core challenges of robotic manipulation within unknown environments under limited sensing capabilities.  First, the force feedback obtained at the robot end effector does not provide direct observation on the hinge (Fig.~\ref{fig:main_figure}) of the Velcro. In fact, we show that one of the state space parameters is unobservable. Second, the unknown geometry adds uncertainties to the Velcro state even with a perfect state estimated in the previous time step. Lastly, we also show that the previously mentioned challenges require the peeling controller to introduce a trade-off between efficiency and state estimation accuracy, which needs to be systematically resolved.

Peeling a Velcro straps involves attaching the gripper to one end of the strap and then pulling it along a trajectory to peel it off. In a general application setting, a sensor such as a camera is likely needed for the attachment phase. In our previous work, we demonstrated that visual input significantly improves the performance of reinforcement learning algorithms for velcro peeling~\cite{yuan2021multi}. However, once the effector is attached, humans can peel off the strap with their eyes closed. In this work, we study this minimal sensing setting and do not consider vision as a modality. Beyond theoretical interest, this setting has practical applications: in a home-care setting the robot may need to operate in low-light or occluded settings relying primarily on tactile and force feedback. 

One of the main technical innovations we introduce in this work is to develop a separate controller that keeps the peeled part of the Velcro taut throughout the peeling task in order to avoid dealing with the ``tactile void space" challenge investigated in~\cite{yuan2021multi}. Then, based on this taut assumption, we model the system state, action, and state transition based on quasi-static dynamics assumptions, providing a tractable representation of the system dynamics. To establish a baseline, we investigate a fully-observable scenario where both the Velcro and robot states are fully known. This allows us to derive a closed-form optimal solution that minimizes energy cost, providing a theoretical performance limit for the task. Building upon this baseline, we address the more realistic and challenging partially-observable case.  We develop a state estimator that leverages state space decomposition to infer the underlying state of the system, including the hidden surface geometry, from noisy force and position feedback. This estimated state information is then used by a heuristics-based controller designed to balance exploration and exploitation, where the robot seeks to gather information about the environment, and it leverages gathered knowledge to efficiently complete the peeling task.

The main contributions of our work are as follows:
\begin{itemize}
    \item This work advances recent research on deformable object manipulation by incorporating frequent forceful interactions with the object. 
    \item We propose a novel method combining analytical modeling, Bayesian state estimation, and heuristics-based control to address the challenges in robot Velcro peeling with force feedback. 
    \item We validate the effectiveness and robustness of our proposed method through extensive experiments featuring complex geometric uncertainties and realistic sensor noise and achieved 100\% success rate with less than 80\% increase in energy cost compared to the optimal solution.

\end{itemize}

\section{Related Work}
In this section, we first review relevant work on robotic manipulation with contact, followed by the progress in active perception.

\subsection{Manipulation with Contact}
Contact in robotic manipulation has been challenging because the nonlinear dynamics of physical contact are difficult to model and predict, especially when friction is involved~\cite{suomalainen2022survey,yuan2023active, leidner2019cognition}. 
In~\cite{laezza2021reform} the authors presented a toolbox for simulating and testing RL algorithms on deformable linear objects. Similarly, recent works in RL focus a lot on tasks like cloth folding~\cite{cloth_petrik2020static, cloth_tsurumine2019deep}. In this work, we push the boundary further by studying a task that incorporates deformable objects with more frequent and stronger forceful interactions. In our previous work~[1], we presented a model-free RL method to address robotic Velcro peeling with force feedback, but there is still a significant performance gap between the RL method and the optimal peeling strategies. In this work, we aim to reduce this gap via a different approach.

Sensing in applications involving contact usually relies on wrist mounted force-torque sensors~\cite{pih_petrik2020static, pih_van2018comparative, pih_zhang2017peg} and tactile sensors~\cite{tactile_chebotar2014learning, tactile_kronander2013learning} on the fingertips of the gripper. However, contact sensing only provides limited information local to the contact regions and ambiguities arise when associating the sensor feedback with the state of areas outside the contact regions. The ambiguities become worse when articulated objects are involved~\cite{articulate_carrera2015learning, articulate_karayiannidis2016adaptive, articulate_tanwani2016learning}. For Velcro straps, the complexity induced by deformation is even higher than doors or jars. However, we present a simple analytic formulation for the Velcro state and let the state estimator handle the deformation uncertainties. Lastly, manipulation with contact is usually more difficult to control robustly and safely. Methods exploring physical systems often alleviate the potential hazard by limiting contact forces using torque control~\cite{levine2015learning} or impedance control with limited stiffness ~\cite{hazara2016reinforcement}. In this work, we design an action cost that takes safety as part of the considerations and show that accuracy in state estimation is also critical to prevent the state from getting close to unwanted regions in the first place.

\subsection{Active Perception}
Active perception~\cite{bohg2017interactive} refers to scenarios in which a robot actively interacts with its environment to improve its perception and understanding. Rather than performing passive observation, the robot strategically plans and performs actions to gather information, thereby reducing uncertainty about the state of the environment. Active perception allows robots to discover hidden properties (e.g., weight distribution~\cite{yuan2023active}, frictional properties, softness/hardness~\cite{natale2004learning}) that are impossible or difficult to measure passively.  Atkeson et al.~\cite{atkeson1986estimation} proposed to move the object to estimate its inertial properties of object dynamics, which are otherwise unobservable. Karayiannidis et al.~\cite{karayiannidis2013model} demonstrated that the properties of articulated objects such as doors and drawers can also be estimated through forceful interactions based on wrist-mounted force-torque sensor without prior knowledge about the joint types, etc. In our case, we show that the Velcro state also contains a hidden parameter, thus requires certain combinations of actions to be estimated accurately. 

Active perception inherently couples perception with action planning and control. The robot must reason explicitly about ``what actions to take," ``when," and ``how," to maximize information gain or reduce uncertainty about its environment. Determining the best actions to reduce uncertainty while optimizing for certain manipulation goals could be computationally challenging, especially for high-dimensional state and action spaces. Authors in~\cite{koval2016pre, javdani2013efficient} demonstrated that Partially Observable Markov Decision Process (POMDP) can be a powerful tool to handle such challenges. Due to the difficulties in estimating the hidden parameter in Velcro state, we build upon the simple yet powerful particle filter to explicitly estimate the state and reason for actions based on the estimation. As for the actions, unlike some prior works~\cite{dragiev2011gaussian, dragiev2013uncertainty} that focus on reasoning low-level control command and generate actions at a high frequency, we adopt a dynamics model based on quasi-static assumptions, to reduce the modeling complexity and ease the requirements for computation speed and hardware latency.
\section{Methodology}

\begin{figure}[t]
\centering
\includegraphics[width=0.36\textwidth]{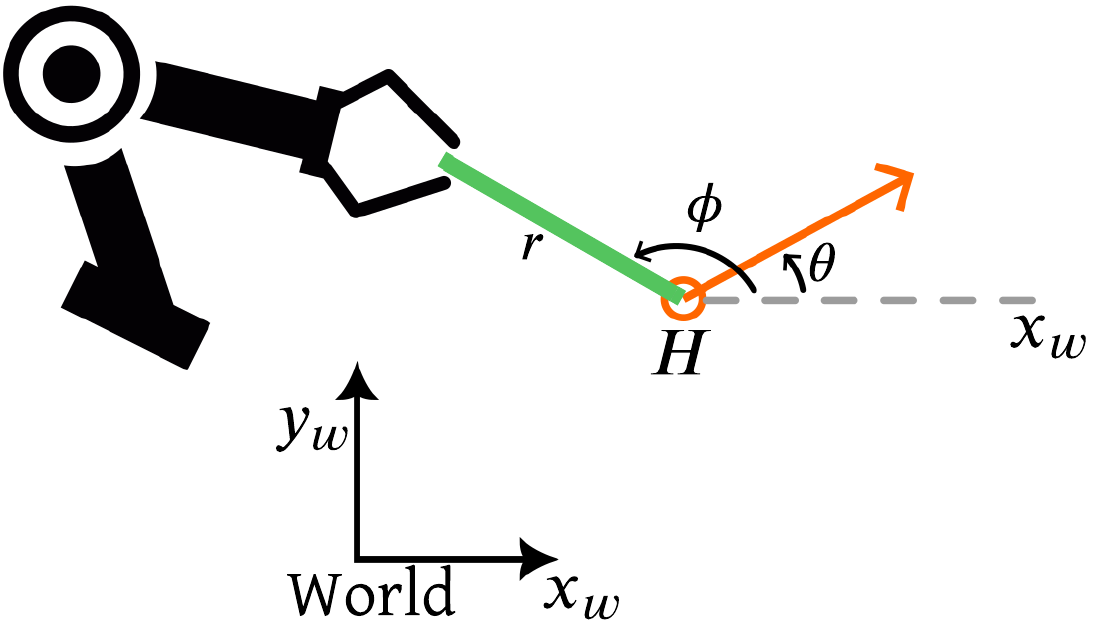}
\caption{\small Illustration of the Velcro peeling state.}
\label{fig:peeling_state}
\end{figure}

This section first introduces the Velcro peeling state model and robot actions with respect to this model. We then explain how we implement the state estimator for this state based on the particle filter. Finally, we discuss the energy cost associated with different actions and present the heuristic approach for the peeling controller.

\subsection{Velcro Modeling}
\label{sec:velcro_modeling}

We assume the peeled part of the Velcro is always taut. In simulation, this means treating the peeled part as a straight line. On a real robot platform, this means the robot must maintain a load on the force sensor that is above a certain threshold. Because of this tautness assumption, the hinge state and the tip position can describe the peeling configuration of a given Velcro at any time step. As illustrated in Fig.~\ref{fig:peeling_state}, $H = [h_x, h_y]$ stand for the position of the hinge in the world frame. The orange vector represents the orientation of the attached part around the hinge, which can then be represented by the angle $\theta$. The peeled part, when in taut, can be represented by angle $\phi$ and length $r$. Therefore, we can write the peeling state $X$ as:

\begin{equation}
\label{eq:state}
    X = \begin{bmatrix}
        h_x \ h_y \ \theta \ \phi \ r
    \end{bmatrix}
\end{equation}

\begin{figure}[t]
    \centering
    \includegraphics[width=0.49\textwidth]{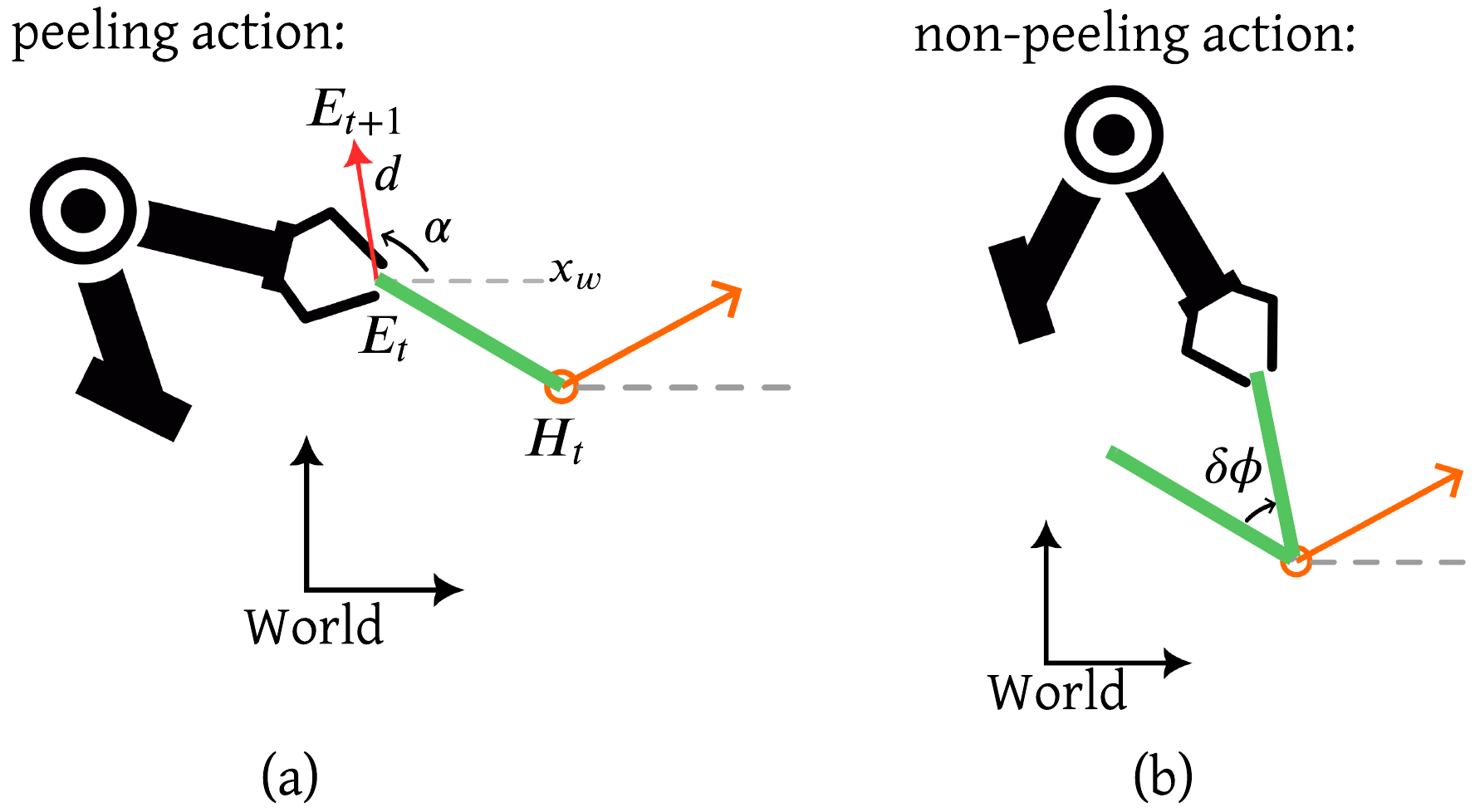}
    \caption{\small Velcro peeling actions.}
    \label{fig:peeling_actions}
\end{figure}

As for the robot actions, the tautness assumption allows us to classify them into two categories: peeling and non-peeling. Intuitively, in both types of actions the end effector moves and maintains the tautness, the Velcro gets peeled more after peeling actions and no peeling happens after non-peeling actions. As shown in Fig.~\ref{fig:peeling_actions}a, a peeling action is defined by the movement of the tip position from point $E_t$ to $E_{t+1}$. The vector $\overline{E_tE_{t+1}}$ can be represented by the angular distance $\alpha$ to x-axis of the world frame and $d$, the distance between point $E_t$ to $E_{t+1}$. For peeling actions, the hinge point $H_t$ will be altered by the action deterministically and we will go over the geometric constraints to solve for $H_{t+1}$ in the remaining of this section. For a non-peeling action, it is essentially changing $\phi$ in the state $X$ while keeping the rest parameters untouched as a result of the taut-peeled-part assumption. To conclude, we can parametrize action $A$ as Eq~\ref{eq:action}, where $s$ is a binary value indicating the category of the action.

\begin{equation}
    \begin{aligned}
        \label{eq:action}
        & A = \begin{bmatrix}
            \alpha \ d \ s \ \delta\phi
        \end{bmatrix}     \text{\ , where \ } \\
        & s = 0 \text{ or } 1 \text{ denotes peeling or non-peeling} 
    \end{aligned}
\end{equation}

\begin{figure}[t]
    \centering
    \includegraphics[width=0.26\textwidth]{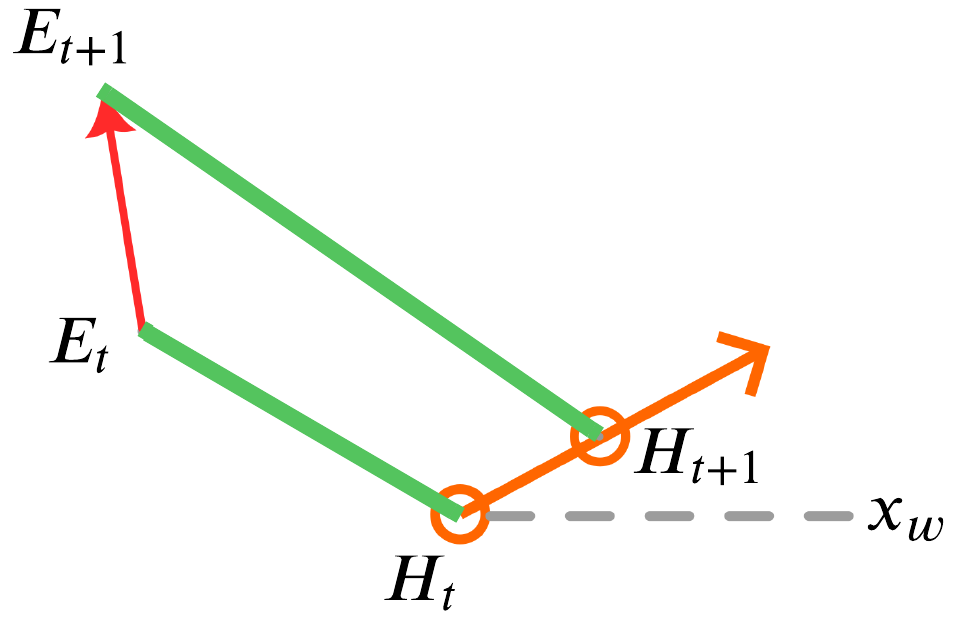}
    \caption{\small Peeling state transition with peeling actions.}
    \label{fig:peeling_actions_transition}
\end{figure}

Given the state and action representations, the state transition of Velcro peeling can also be derived based on quasi-static dynamics assumptions. In the case of peeling actions, the state and action at time $t$ is $X_t = [h_{x,t} \ h_{y,t} \ \theta_t \ \phi_t \ r_t]$ and $A_t = [\alpha\ d\ 0 \ 0]$. The next state must satisfy the geometric constraints illustrated in Fig.~\ref{fig:peeling_actions_transition}. The sum of $\Vert E_t H_t \Vert$ (the length of the peeled part at time $t$) and $ \Vert H_tH_{t+1} \Vert$ (the length of the peeled Velcro), should equal to $\Vert E_{t+1}H_{t+1} \Vert $ (the length of the peeled part at time $t+1$). Note that the equation still holds even if the underlying geometry is not flat, only the length of the peeled Velcro might no longer be $ \Vert H_tH_{t+1} \Vert$. Let the state at time $t+1$ be $X_{t+1} = [h_{x,t+1} \ h_{y,t+1} \ \theta_{t+1} \ \phi_{t+1} \ r_{t+1}]$ and $dr = r_{t+1} - r_t$. For a peeling action with arbitrarily small $d$, the local surface of the underlying geometry can be approximated with a straight vector, therefore the constraint for the transition model can be written as:

\begin{equation}\label{eq:geometric_constraint}
    \begin{cases}
        &r_t \cos{\phi_t} + d \cos{\alpha_t} = dr \cos{\theta_t} + r_{t+1} \cos{\phi_{t+1}} \\
        &r_t \sin{\phi_t} + d \sin{\alpha_t} = dr \sin{\theta_t} + r_{t+1} \sin{\phi_{t+1}} \\
        &h_{x, t+1} = h_{x, t} + dr \cos{\theta_t} \\
        &h_{y, t+1} = h_{y, t} + dr \sin{\theta_t}
    \end{cases}
\end{equation}

Essentially, after action $A_t$ determines the relative position between $E_{t+1}$ and $E_t$, solving Eq.~\ref{eq:geometric_constraint} yields the relative position between $H_t$ and $H_{t+1}$. However, it is worth noting that $\theta_{t+1}$ does not appear in Eq.~\ref{eq:geometric_constraint}. The rationale behind the unconstrained $\theta_{t+1}$ is that the orientation of the underlying surface depends neither on the previous states nor previous actions. In other words, if we apply the Markovian assumption, where the current state is only dependent on the state and action from the previous time step, then $h_x, h_y,r,\phi$ are observable parameters but $\theta$ is unobservable. This property turns out to be one of the main challenges for Velcro state estimation which we are going to introduce in the following section.

Finally, we introduce the observation model in the process of Velcro peeling. We assume that the end effector orientation is irrelevant and therefore the effector can be treated as a point. Therefore, along with a force sensor, the robot agent will observe the tip position of the Velcro as well as a force vector due to the tension in the peeled part. Because we adopt the quasi-static dynamics model, the magnitude of force does not matter. The observation vector $O$ in Eq.~\ref{eq:obs} consists of the tip position $t_x, t_y$ and angle $\beta$ denoting the direction of the force vector.

\begin{equation}
\label{eq:obs}
    O = \begin{bmatrix}
        t_x \ t_y \ \beta
    \end{bmatrix}
\end{equation}

\subsection{Velcro State Estimation}
\label{sec:state_estimation}
To estimate the Velcro state $X$, we present a particle filter with state space decomposition due to the discrete and nonlinear properties of the dynamics model. Depending on the action category and the measurement type, the particle filter selectively updates a subset of the state space parameters to avoid interference with the distribution of the unobserved state space parameters.

We mostly follow the filter design in~\cite{gustafsson2002particle} with our state, action, and transition model except for the importance sampling and the state space decomposition. In total, we present three different measurement models in Eq.\ref{eq:measure_beta},~\ref{eq:measure_tip},~\ref{eq:measure_theta} to update the weight of the particles for the resampling step. Each of them corresponds to a different subset of the state space parameters as follows.

\begin{align}
     \beta \ &= \phi + \pi + \mathcal{N}(0,\,\sigma_1^{2}) \label{eq:measure_beta} \\
    \begin{bmatrix} t_x \\ t_y \end{bmatrix} &= 
      \begin{bmatrix} h_x \\ h_y \end{bmatrix} + r * \begin{bmatrix} \cos{\phi} \\ \sin{\phi} \end{bmatrix} + \mathcal{N}(\textbf{0},\,\sigma_2^{2}) \label{eq:measure_tip} \\
    \theta_{t+1} &= \theta_t + \mathcal{N}(0,\,\sigma_3^{2}) \label{eq:measure_theta}
\end{align}

We use the classical design of the update scheme for the particle filter similar to Gustafsson et al.\ \cite{gustafsson2002particle} where $P(O_t|X_t)$ is assumed to be a Gaussian distribution. In Eq.~\ref{eq:measure_beta}, for an unbiased force sensor, the measured force vector should be pointed toward the opposite direction of the angle $\phi$ with a zero-mean Gaussian noise $\mathcal{N}(0,\,\sigma_1^{2})$. In Eq.~\ref{eq:measure_tip}, we use another Gaussian $\mathcal{N}(\textbf{0},\,\sigma_2^{2})$ to model the residual between the measured tip position and the tip position based on the state parameters $h_x, h_y, r, \phi$.
Finally, as we have shown in Eq.~\ref{eq:geometric_constraint}, if we don't adopt any assumptions on the geometric shape of the underlying surface, the current $\theta$ is not constrained by the previous states, meaning the surface can be arbitrarily uneven. To apply this method to real-world Velcro peeling applications, we argue that imposing a certain level of radius constraints on the underlying surface is reasonable. Therefore, in Eq.~\ref{eq:measure_theta} we model the residual between $\theta_{t+1}$ and $\theta_t$ to estimate $\theta$. However, the shape of the surfaces that the Velcro is often applied to in the real world, including circles and corners, do not always fit in the zero-mean Gaussian model. As a result, we use an auxiliary state estimation model specifically designed for $\theta$ to first reduce the bias of $\theta_t$ in Eq.~\ref{eq:measure_theta} and then apply the zero-mean Gaussian model.

The auxiliary state estimation model for $\theta$ is a combination of a line-fitting algorithm based on Principal component analysis (PCA) and a least-square-based arc-fitting algorithm. The hinge positions estimated from previous steps act as the input for the fitting and we apply a decayed weight function to marginalize the influence of samples over time. The results from both fitting algorithms are then selected based on the fitting covariance. 

When the action is a peeling action, all the parameters in the state $X$ are coupled together but our model does not consider the covariance matrix as a whole. Updating all of them turned out to increase both the bias and the variance of the particles. So in this case, the filter only uses Eq.~\ref{eq:measure_beta} in the update step and updates $\phi$. For non-peeling actions, the hinge and the peeled length stay constant, meaning $h_x, h_y, r$ are decoupled with $\phi$ so we use Eq.~\ref{eq:measure_beta} and \ref{eq:measure_tip} to update the particles. Finally, an update on $\theta$ can be done at any step and we choose to let the controller decide when it is needed. 
In the remaining sections, we use $\mathscr{F}$ to denote the particle filter and use $\mathscr{F}_1, \mathscr{F}_2, \mathscr{F}_3$ to refer to the update steps with measurement models introduced in Eq.~\ref{eq:measure_beta} \ref{eq:measure_tip} \ref{eq:measure_theta}.

\begin{algorithm}[h]
    \caption{Velcro Peeling Controller}
    \label{alg:peeling}
    \SetAlgoLined
    Randomly initialize particles \textbf{P} in $\mathscr{F}$\\
    \While{peeling has not finished}{
        \uIf{$\phi - \theta \neq \pi/2$}{
            Set $A = \begin{bmatrix} 0, 0, 1, \pi/2 -(\phi - \theta) \end{bmatrix}$ \\
            Execute $A$ and step $\mathscr{F}_1$ and $\mathscr{F}_2$ 
        }

        Set $A = \begin{bmatrix} \theta + \pi / 4, 1, 0, 0 \end{bmatrix}$\\
        Execute $A$ and step $\mathscr{F}_1$
        
        Compute the health index $z$
        
        $s = \begin{cases} 
          1, & \text{if } \mathbf{rand()} < z \\
          0, & \text{otherwise} \end{cases}$
        
        \uIf{$s = 1$}{
            Set $A = \begin{bmatrix} 0, 0, 1, - \pi/4 \end{bmatrix}$ \\
            Execute $A$ and step $\mathscr{F}_1$ and $\mathscr{F}_2$ \\
            Run auxiliary estimator on $\theta$ \\
            Step $\mathscr{F}_3$
        }
    }
\end{algorithm}

\subsection{Velcro Peeling Controller}
\label{sec:peeling_controller}

Finally, we will go over the definition of the energy cost associated with Velcro peeling and how we design the heuristics-based controller to achieve robust and efficient peeling. 

We take inspiration from human peeling behaviors when we define the peeling cost. We observe that humans tend to keep the peeled part perpendicular to the underlying surface throughout the peeling process. There are a few benefits of such a configuration. It is less likely for the end effector to collide with the underlying surface. It also stays away from situations where the peeled part wraps on top of the attached part, causing entanglement, or peeling towards or close to the opposite direction of $\theta$, which will also block the peeling process. As a result, we design a potential function in Eq.~\ref{eq:potential} that tracks the angle between the peeled and attached part and adds quadratic penalization to the deviation from the right angle. Then, the cost in Eq.~\ref{eq:cost}, can be defined by integrating the potential along a path. For peeling actions, the path goes from $r_t$ to $r_{t+1}$ and for non-peeling actions, it goes from $\phi_t$ to $\phi_{t+1}$.

\begin{align}
    U(X) &= 1 + \Vert \phi - \theta - \pi/2 \Vert^{2} \label{eq:potential} \\
    C(X_t, A_t, X_{t+1}) & =  
    \begin{cases}
         c_1 \int_{r_t}^{r_{t+1}} U(X) \,dr  &\mbox{if } s_t = 0 \\
         c_2 \int_{\phi_t}^{\phi_{t+1}} U(X) \,d\phi   &\mbox{if } s_t = 1 
    \end{cases} \label{eq:cost} 
\end{align}

The goal of the controller is to peel open the whole Velcro strap while minimizing the total energy cost.
Meanwhile, given the properties of the state, action, transition model, and state estimator that have been discussed in previous sections, it is clear that maintaining a high-quality state estimation is crucial throughout the process, and the controller should balance exploratory (non-peeling action) and exploitative (peeling action) behaviors. The details for the peeling controller are presented in Algorithm~\ref{alg:peeling}. One interesting behavior of the peeling action on flat Velcros is that, $\phi$ will remain the same when $\alpha = \theta + (\phi-\theta)/2$ throughout the action. We leverage this property to design the heuristics such that, a non-peeling action first brings the peeled part to the right angle and $\alpha = \theta + \pi/4 $ for the peeling action in every iteration. After the particle filter update step after the peeling action, the distribution of the likelihood for the particles can provide some information to measure the risk of sample impoverishment. Specifically, if the distribution shifts more towards $0$, this indicates that less particles are close to the true state. In this case, the controller should adopt more non-peeling actions to reduce the variance of the observable state parameters, or add more process noise (roughening~\cite{gordon1993novel}) to $\theta$. In practice, we use a health index $z$ to denote the likelihood of sample impoverishment and it is equal to the ratio of particles with weight less than 0.1 minus the ratio of particles with weight larger than 0.9. It is then clipped on the left side by 0 to serve as the threshold to determine if we need an additional non-peeling action before the next iteration.

\section{Evaluation and Results}
In this section, we first introduce the experiments we designed to evaluate the performance of Velcro peeling controllers. Next, we introduce the evaluation metrics, and the baseline methods that are used to compare with our approach. Finally, we present the results of all the methods.

\begin{figure}
    \centering
    \begin{subfigure}{0.15\textwidth}
        \includegraphics[width=\textwidth]{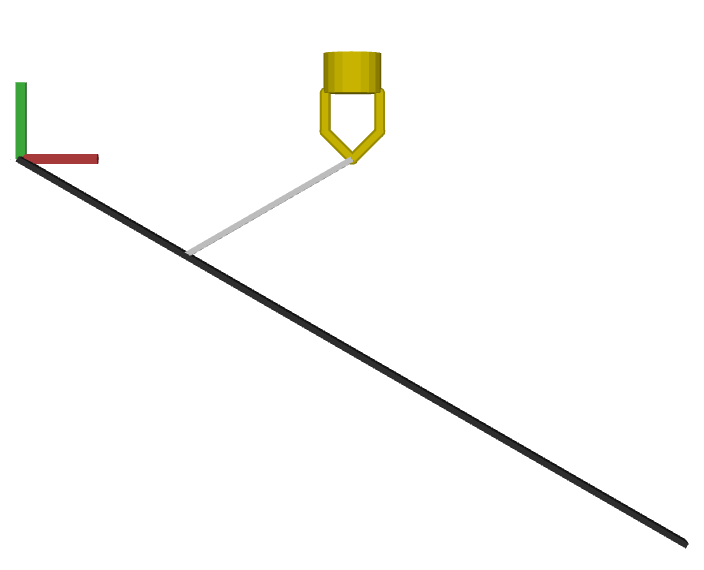}
        \caption{\small Flat Velcro.}
        \label{fig:flat}
    \end{subfigure}
    \hfill
    \begin{subfigure}{0.15\textwidth}
        \includegraphics[width=\textwidth]{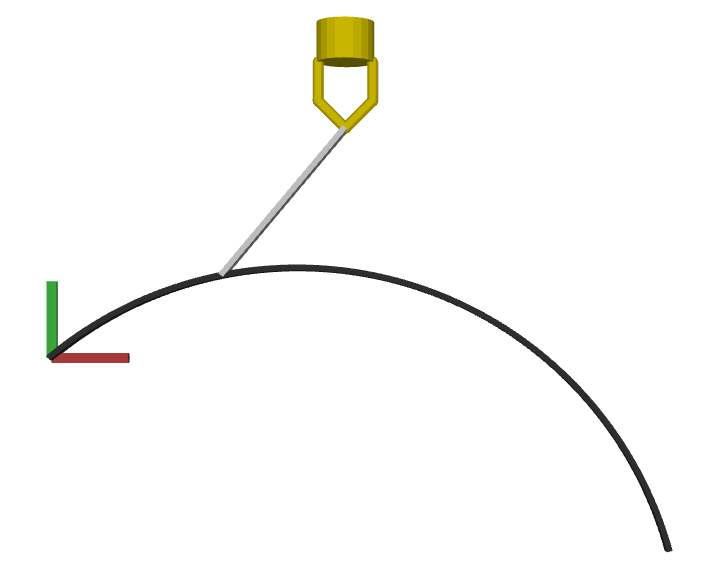}
        \caption{\small Arc Velcro.}
        \label{fig:arc}
    \end{subfigure}
    \hfill
    \begin{subfigure}{0.15\textwidth}
        \includegraphics[width=\textwidth]{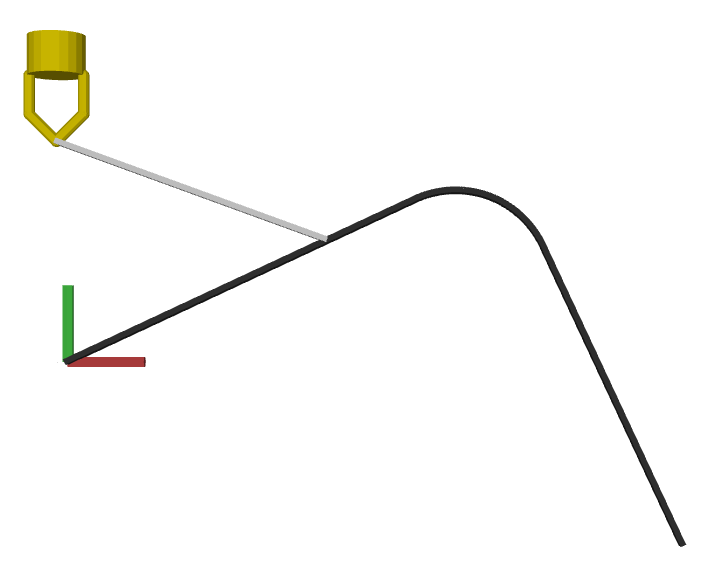}
        \caption{\small Corner Velcro.}
        \label{fig:corner}
    \end{subfigure}
            
    \caption{\small{Examples of various Velcro shapes and poses generated for evaluation with the underlying surface (black), the gripper (yellow), the peeled part (gray), and a coordinate frame (red/green) centered at the starting point of the Velcro. (a) Flat Velcro with a negative tilt angle. (b) Arc Velcro with a positive tilt. (c) Corner Velcro with a positive tilt angle and a rounded corner.}}
    \label{fig:velcro shapes}
\end{figure}

\subsection{Experiments}
\label{sec:experiments}

To evaluate our method, we generate Velcro strips with various shapes and poses including flat, arc, and corner as illustrated in Fig.~\ref{fig:velcro shapes}. Every strip has a fixed length of 60 centimeters and the peeling initializes with a peeled length of 10. For the flat case in Fig.~\ref{fig:flat}, we only vary the tilt angle of the surface between $-60^{\circ}$ and $60^{\circ}$. For the arc shape in Fig.~\ref{fig:arc}, the radius is randomly sampled between $20$ and $40$ in addition to the random tilt angle. Lastly, for the corner shape in Fig.~\ref{fig:corner}, we adopt a rounded corner with radius randomly sampled between $4$ and $15$ and the ratio of the flat part after the corner is varied between $0.3$ and $0.7$. 

We sample 200 configurations uniformly at random for each of the Velcro shapes. The world frame is set to coincide with the reference frame at the starting point (Fig.~\ref{fig:velcro shapes}). We initialize the Velcro state $X$ based on a fixed peeled length of 10 and $\phi=90^{\circ}$. In simulation, we also add a noise of $\mathcal{N}(0,\,1)$ to $\beta$ to account for sim-to-real gap.

\begin{table}[t!]
    \centering
    \caption{\small{Total Cost ($E$) of Each Episode}}
    \label{tab:cost}
    \begin{tabular}{l || c | c c c} 
        \toprule
        {} & \bf{Full-Obs} & \bf{Ours} & \bf{PF+RL} & \bf{RL} \\
        \midrule            
        Flat   &   50    &  \bf{67.4}  & 122.9  &  143.5  \\
        Arc    &  51.3   &  \bf{85.1}  & 241.6  &  363.9  \\
        Corner &  50.6   &  \bf{103.7}  & 202.7  &  312.1  \\
        \bottomrule
    \end{tabular}
\end{table}

\subsection{Evaluation Metrics}
\label{sec:metrics}

We use $E$ to denote the energy cost from the actions aggregated throughout the peeling process. Besides $E$, we also keep track of the validity of the peeling outcome. For example, if the angle $\phi-\theta$ after the peeling action is less than $5^{\circ}$ or lager than $175^{\circ}$, then it is considered in the forbidden zone, and the peeling episode will be marked as failed and terminated immediately.
We use $\eta$ to denote the success rate of the peeling process.

\begin{figure*}[t]
    \centering
    \includegraphics[width=0.95\textwidth]{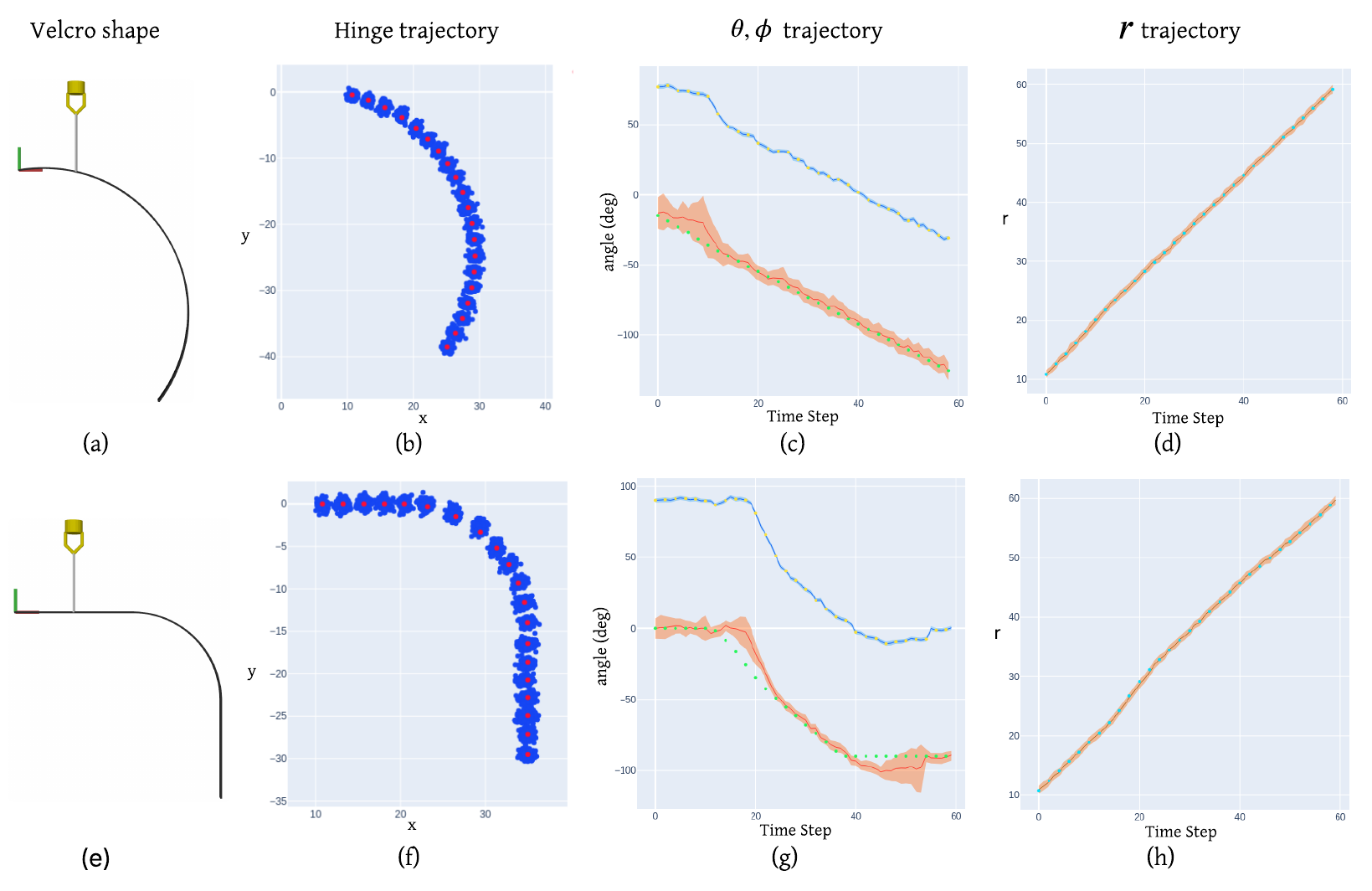}
    \caption{\small Examples of how the particle filter evolves during an episode of Velcro peeling. The first column (a, e) illustrates the shape of an arc Velcro and a corner Velcro. The second column (b, f) plots the hinge trajectory of sim (red) and all the particles (blue) for only a subset of time steps. The third column (c, g) corresponds to the mean and spread of $\theta$ (orange) and $\phi$ (blue) for particles with dotted markers for sim state. (d, h)  Mean and spread of $r$ (orange) for particles with dotted markers for sim state.}
    \label{fig:results}
\end{figure*}

\subsection{Baselines}
\label{sec:baselines}

To provide some references for the performance of our method, we present three different baseline methods as follows:

\textbf{Full-Obs} One of the main challenges in Velcro peeling with only force feedback is the partial observability of the Velcro state. If the state is fully observable, a simple heuristics can be designed to keep the trajectory as close to the minimum potential region as possible. The aggregated cost in this case may act as the lower bound for the partially observable case. To ensure a fair comparison, we also use fixed length straight line actions similar to the actions used in Algorithm.~\ref{alg:peeling}.

\textbf{PF+RL}
To demonstrate the efficiency and robustness of our controller, we implemented a baseline that combines our particle filter state estimator with a model-free RL based agent. We call this approach PF+RL. In this method, we use the weighted average state from the particle filter as the state vector for the RL agent. We use Proximal Policy Optimization (PPO)~\cite{schulman2017proximal} algorithm with the Transformer architecture~\cite{vaswani2017attention} to train the controller. The goal is to leverage the power of the Transformer's attention mechanism and the stability of PPO to train an agent that can be both robust and efficient. The negative energy cost becomes the reward target to maximize for. As for the actions, we adopt a continuous action space to sample based on the mean and variance output by the policy network.

\textbf{RL}
The pure RL method is based on the POMDP assumption similar to~\cite{yuan2021multi}. The main difference between RL method and PF+RL method is the utilization of a state estimator. Instead of explicitly inferencing the state vector based on the history of observations, we use the Transformer network to directly encode the observations and output the actions. The reward and action space design is identical to the PF+RL method.

\subsection{Results}
\label{sec:results}

\begin{table}[t!]
    \centering
    \caption{\small{Success Rate ($\bar\eta$ / $\%$) of Each Episode}}
    \label{tab:eta}
    \begin{tabular}{l || c | c c c} 
        \toprule
        {} & \bf{Full-Obs} & \bf{Ours} & \bf{PF+RL} & \bf{RL} \\
        \midrule                  
        Flat   &  100   &  \bf{100}  & 97.5  &  90  \\
        Arc    &  100   &  \bf{100}  & 81    &  83  \\
        Corner &  100   &  \bf{98.5}  & 61  &  45.5  \\
        \bottomrule
    \end{tabular}
\end{table}

In Table.~\ref{tab:cost} and Table.~\ref{tab:eta}, we present the total cost $E$ in each episode and the success rate $\eta$ for all the baseline methods in each evaluation cases. Overall, flat Velcro cases are the easiest. Both arc and corner Velcros can be challenging. Compared with the other two baseline methods with partial observability, our method can achieve the same level of performance as the fully observable method in terms of success rate with less than $80\%$ increase in energy cost. In comparison, \textbf{PF+RL} method sometimes fails to allocate enough non-peeling actions to help with the particle filter convergence, causing the filter to either diverge or run into sample impoverishment issues frequently. Although POMDP could address the partial observability challenge in theory, we find that the \textbf{RL} method does not perform as well as expected. The reason could be either the drastic changes of the effects to the state estimation process when switching between peeling vs non-peeling actions, or the mixture of highly observable and hidden parameters in the state vector, which contribute to the success of our particle filter using state space decomposition strategy.

In Fig.~\ref{fig:results}, we show two examples of how the particle filter evolves through the peeling process for arc and corner Velcro peeling. For the arc Velcro we set the radius to be 25 and the corner Velcro has a radius of 15 centimeters. For better visualization, we only plot the particles associated with the peeling actions. The hinge trajectory of the simulation state and particles are plotted with red and blue points in Fig.~\ref{fig:results}b and Fig.~\ref{fig:results}f. For $\theta, \phi$ and $r$ we show how the mean and spread evolve over time. In conclusion, the particle filter shows robustness against the changing underlying geometries. The difference between $\phi$ and $\theta$ is also maintained around $\pi/2$, which tells that the controller can keep the state close to the region with low energy cost.

Overall, the results establish that our approach can accurately estimate the Velcro state while finishing the peeling process efficiently with a high success rate.
\section{Conclusions}

Our work advances recent research on deformable object manipulation with force interaction by presenting a novel method to overcome the environment uncertainties and sensor incompleteness in the Velcro peeling task. We demonstrate the capabilities of our proposed method to achieve robust and efficient Velcro peeling by studying the theoretical limits and comparing it with multiple baselines through comprehensive experiment evaluations. We will further highlight the potential of application of our method to the real environment by evaluating the sim-to-real gap of the sensor measurement in the submitted video.
Future directions include exploring  more efficient controllers and incorporating our method into a complete system. 

\bibliographystyle{IEEEtran}
\bibliography{biblio}

\end{document}